# A Fuzzy Inference System for the Identification

J. de J. Rubio, *Member, IEEE*, R. S. Ortigoza, F. J. Avila, A. Melendez and J. M. Stein

*Abstract*— Odor identification is an important area in a wide range of industries like cosmetics, food, beverages and medical diagnosis among others. Odor detection could be done through an array of gas sensors conformed as an electronic nose where a data acquisition module converts sensor signals to a standard output to be analyzed. To facilitate odors detection a system is required for the identification. This paper presents the results of an automated odor identification process implemented by a fuzzy system and an electronic nose. First, an electronic nose prototype is manufactured to detect organic compounds vapor using an array of five tin dioxide gas sensors, an arduino uno board is used as a data acquisition section. Second, an intelligent module with a fuzzy system is considered for the identification of the signals received by the electronic nose. This solution proposes a system to identify odors by using a personal computer. Results show an acceptable precision.

## I. Introducción

LOS HUMANOS emplean los sentidos del gusto, olfato, o tacto para percibir su medio ambiente. En particular, el olfato detecta y clasifica algunos tipos de frutas como son la papaya, la naranja, la manzana, la guayaba, o el plátano, o de vegetales como es la cebolla, algunos olores son placenteros, pero otros olores son desagradables o peligrosos; por lo tanto, se podría usar una nariz electrónica para estos casos.

Existen algunos trabajos que consideran las narices electrónicas. En [1], los autores proponen un modelo de percepción artificial de un sensor de fusión de datos para analizar la información para determinar la calidad del té. En [6], los autores usan un sensor de nariz electrónica para comparar la exactitud de la clasificación para cuatro algoritmos de reconocimiento de patrones. En [7], se presentan seis alcances para el reconocimiento de cuatro jugos de tomate. En [8], se considera la aplicación de un sistema de inferencia neuro-difuso para la estimación de la concentración de compuestos orgánicos. En [9], dos métodos novedosos se usan para evaluar si la combinación de la lengua y la nariz son más efectivas para predecir el tipo de arroz. En [12], se usa una nariz electrónica para mostrar que ciertos alientos tienen algunas características individuales. En [15], se propone un sensor de selección que usa el reconocimiento de patrones en una nariz electrónica. De los resultados anteriores, las narices electrónicas se pueden aplicar como los sensores para la clasificación de objetos por medio de sus olores.

Existen algunos trabajos acerca de lógica difusa. El estudio de [2] describe el clasificador difuso que es usado para el diagnóstico de patologías de la columna vertebral. En [3], se investiga la aplicación de los conjuntos difusos para seleccionar los tamaños óptimos de los circuitos integrados. En [4], se usa el control difuso con la intención de encontrar la frecuencia para manejar un motor de inducción. En [5], se propone un esquema de control difuso que regula o ajusta la posición rotacional de un manipulador. En [10], el objetivo principal es hacer un estudio comparativo del desempeño de cinco técnicas en base a su capacidad de clasificación de datos históricos. En [11], se define un vector regresor adecuado para un sistema y se maneja su implementación a través de un enfoque de caja negra. En [14], se propone una metodología con lógica difusa para estimar los datos en una mina. Los trabajos anteriores muestran que la lógica difusa actualmente tiene una gran cantidad de aplicaciones.

En este estudio, se introduce un método novedoso para la clasificación de frutas o vegetales. Primero, se diseña una nariz electrónica para obtener los datos reales de los olores de los tres objetos: la papaya, la naranja, y la manzana para el ejemplo 1, o la cebolla, la guayaba, y el plátano para el ejemplo 2. Después, se introduce un método de clasificación para detectar entre los tres tipos de objetos, ésta estrategia está basada en un sistema de inferencia difuso.

El resto de este trabajo se organiza como sigue. En la sección II, se describe la nariz electrónica. En la sección III, se muestran las concentraciones de los objetos. En la sección IV, se describe el sistema de inferencia difuso. En la sección V, se describe el algoritmo inteligente por pasos. En la sección VI, se explica en detalle el método de clasificación. En la sección VII, se prueba el método de clasificación propuesto para detectar tres tipos de objetos. Finalmente, en la sección VIII se explican los resultados y la posible investigación futura.

## II. Nariz electrónica

En esta investigación se manufacturó una nariz electrónica con un arreglo de 5 sensores de gas que permite clasificar el aroma de tres objetos. Con el arreglo de sensores se hace posible detectar una gran cantidad de olores dada la sensibilidad a los diferentes compuestos químicos de cada uno de los sensores.

*A. Sensores*

En esta sub-sección, se describen los sensores utilizados en la nariz electrónica. La Tabla I muestra la lista de los sensores de

J. de J. Rubio, Sección de Estudios de Posgrado e Investigación, ESIME Azcapotzalco, Instituto Politécnico Nacional, México, jrubioa@ipn.mx, rubio.josedejesus@gmail.com

R. S. Ortigoza, Departamento de Mecatrónica, CIDETEC, Instituto Politécnico Nacional, México, rsilvao@ipn.mx

F. J. Ávila, Tecnológico de Estudios Superiores de Ecatepec, México, favila11@udavinci.edu.mx

A. Meléndez, Tecnológico de Estudios Superiores de Ecatepec, México, amelendez11@udavinci.edu.mx

J. M. Stein, Tecnológico de Estudios Superiores de Ecatepec, México, jstein11@udavinci.edu.mx

gas utilizados en la construcción del arreglo de sensores de la nariz electrónica utilizada en esta investigación.

TABLA I. SENSORES DE GAS DE LA NARIZ ELECTRÓNICA.

| Sensor | Compuestos químicos |
|---|---|
| $MQ-2$ | $H_2$, LPG, $CH_4$, CO |
| $MQ-3$ | Alcohol, Benceno, $CH_4$, Hexano |
| $MQ-135$ | $CO_2$, $NH_3$, $NO_x$, CO, Alcohol, Benceno |
| $TGS2610$ | LPG, Metano, Hidrogeno, Isobutano, Propano |
| $TGS2611$ | Metano, Isobutano, Metanol |

El MQ-2 es un sensor de gas que está compuesto por un micro tubo cerámico $AL_2O_3$, Dióxido de Estaño ($SnO_2$) de capa sensible, el electrodo de medición y el calentador se fijan en una corteza hecha por plástico y malla de acero inoxidable. El calentador ofrece condiciones de trabajo necesarias para el trabajo de los componentes sensibles. El MQ-2 con envoltura tiene 6 pines, 4 de ellos se utilizan para buscar señales y los otros 2 se utilizan para proporcionar la corriente de calentamiento. Este sensor obtiene las concentraciones de los gases $H_2$, $C_3H_8$, $C_4H_{10}$.

El material sensible del sensor de gas MQ-3 es el $SnO_2$, con poca conductividad en el aire limpio. Pero cuando existe el gas alcohol, la conductividad del sensor es más alta, junto con la concentración de gas que va en aumento. Este sensor obtiene las concentraciones de los gases $C_6H_6$, $C_2H_{60}$.

El MQ-135 se utiliza en equipos de control de calidad del aire para edificios y oficinas, es adecuado para la detección de $NH_3$, $NO_X$, alcohol, benceno, humo, $CO_2$, etc. Este sensor obtiene las concentraciones de los gases $CO_2$, $NH_3$, $N_2$.

El TGS-2610 es un sensor de gas semiconductor que combina muy alta sensibilidad al gas licuado de petróleo (LP) con bajo consumo de energía y larga vida. Este sensor obtiene las concentraciones de los gases $C_3H_8$, $C_4H_{10}$, $CH_4$, $H_2$, $C_2H_{60}$.

El TGS-2611 es un sensor de gas semiconductor que combina muy alta sensibilidad al gas metano con bajo consumo de energía y larga vida. Este sensor obtiene la concentración del gas $CH_4$.

Estos sensores fueron colocados en una placa para conformar el arreglo con la configuración electrónica de la Fig. 1.

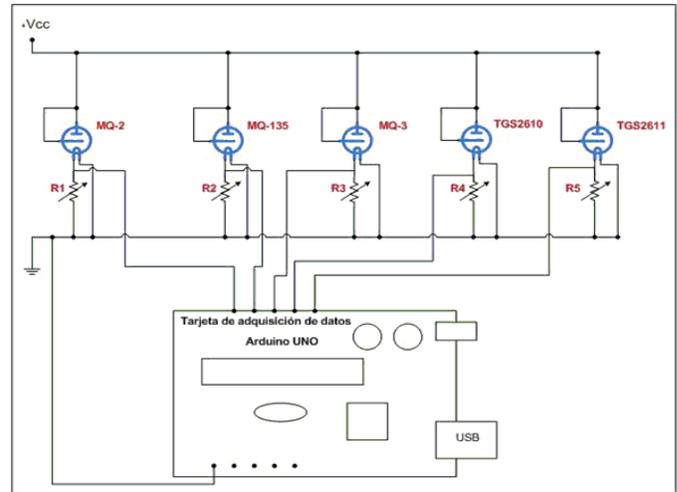

Figura 1. Diagrama de la nariz electrónica.

Cada sensor se acopla con una resistencia variable o potenciómetro en el pin del electrodo para crear un voltaje de salida. Esta resistencia se ajusta para establecer la sensibilidad del sensor a los diferentes compuestos o gases que se pretenden detectar.

El voltaje de salida en la resistencia de ajuste de cada sensor varía entre 0 y 5 volts y se conduce a la tarjeta de adquisición de datos (arduino uno) con el fin de obtener el valor de cada sensor en presencia de uno de las objetos. Los 5 sensores son procesados en el arduino para registrar todas las variaciones y construir una tabla de datos en la computadora concernientes a cada una de los objetos.

En el arduino los voltajes obtenidos en las entradas analógicas son amplificados y mapeados a valores enteros que van de 0 a 1023, lo que representa un valor de 4.9 mV por unidad, en este dispositivo se toma un tiempo de 100 micro segundos en leer las entradas analógicas.

*B. Microcontrolador*

Arduino es el dispositivo usado en este artículo, primero, para amplificar las señales obtenidas por los sensores, y segundo, para enviar los datos desde la etapa de amplificación a una computadora personal. La Fig. 2 muestra el microcontrolador arduino.

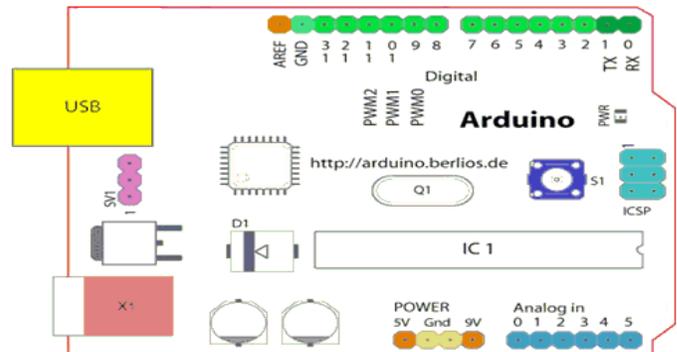

Figura 2. Microcontrolador Arduino.

### III. Concentración de los objetos

Los valores obtenidos en el proceso de detección de aromas representan las 5 entradas para el algoritmo de aprendizaje y se denotan como $z_1(k)$=MQ-135, $z_2(k)$=TGS-2610, $z_3(k)$=MQ-2, $z_4(k)$=TGS-2611, $z_5(k)$=MQ-3, ellas son diferentes para los tres objetos.

La Fig. 3, Fig. 4, y Fig. 5 muestran las señales electrónicas de las concentraciones de los tres objetos utilizados para el ejemplo 1: la papaya, la naranja, y de la manzana.

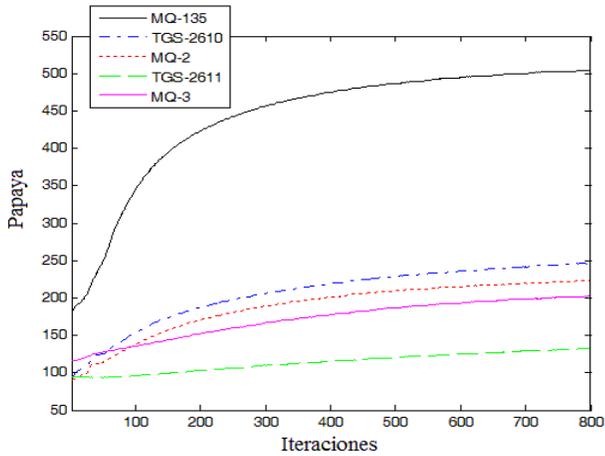
Figura 3. Concentraciones para la papaya.

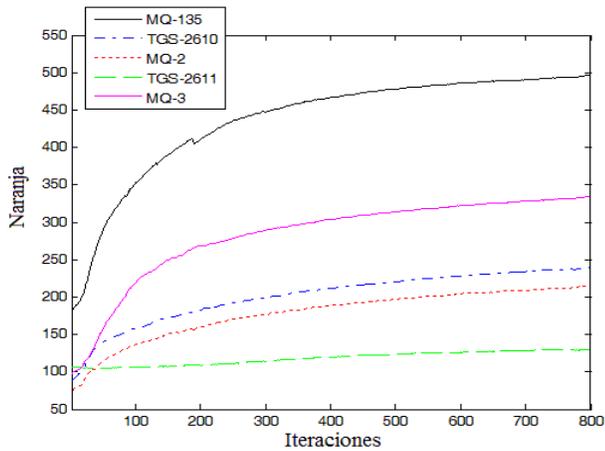
Figura 4. Concentraciones para la naranja.

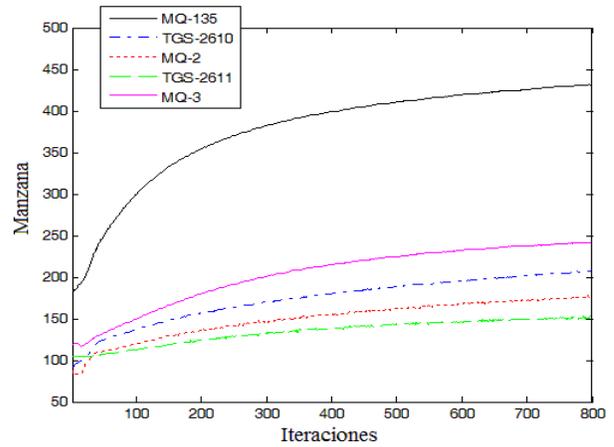
Figura 5. Concentraciones para la manzana.

La Fig. 6, Fig. 7, y Fig. 8 muestran las señales electrónicas de las concentraciones de los tres objetos usados para el ejemplo 2: la cebolla, la guayaba, y el plátano.

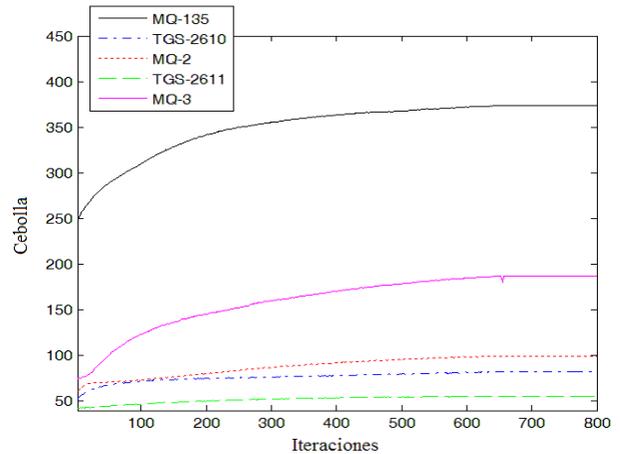
Figura 6. Concentraciones para la cebolla.

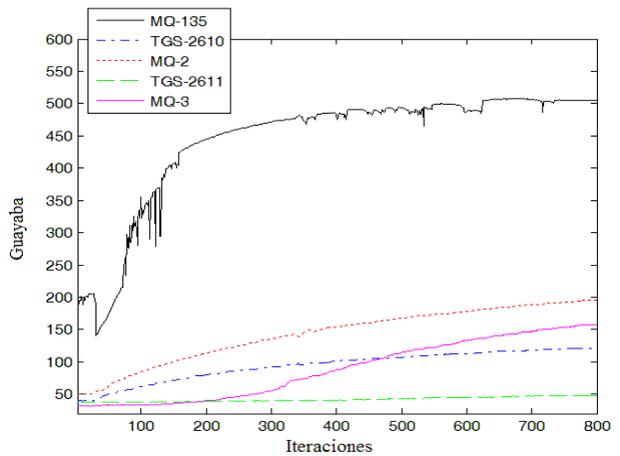
Figura 7. Concentraciones para la guayaba.

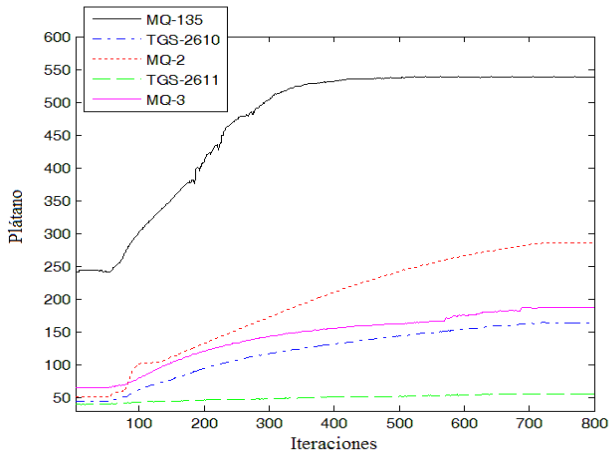

Figura 8. Concentraciones para el plátano.

Para el registro de los datos se establece un proceso de detección de olores con la nariz electrónica para tomar las muestras de los 3 objetos.

Inicialmente se dejó a la nariz electrónica en reposo durante 10 minutos para después incorporar el objeto 1, y tomar las lecturas durante 3 minutos, posteriormente se liberó el aroma del objeto dejando al dispositivo en reposo durante 3 minutos. Una vez transcurrido este tiempo se incorporó el objeto 2 y se tomaron las muestras, se continuó liberando el aroma del objeto durante 3 minutos. Y se finalizó incorporando el objeto 3 en la nariz electrónica para tomar los registros de datos durante 3 minutos. Esto permitió obtener alrededor de 800 datos para cada objeto.

En el rango de los 800 datos obtenidos para cada objeto, los valores varían de acuerdo a la concentración del olor que aumenta conforme transcurre el tiempo.

## IV. ALGORITMO INTELIGENTE

En esta sección, se presenta el algoritmo de lógica difusa que será usado por el método propuesto para la clasificación de datos. El algoritmo es el sistema de inferencia difuso. Las cinco entradas del algoritmo inteligente son $z_1(k)$=MQ-135, $z_2(k)$=TGS-2610, $z_3(k)$=MQ-2, $z_4(k)$=TGS-2611, $z_5(k)$=MQ-3, los cuales tienen valores diferentes para las concentraciones de los tres objetos. Las salidas para el algoritmo inteligente son como sigue:

$$y_1(k) = 1$$
$$y_2(k) = 0 \quad (1)$$
$$y_3(k) = 0$$

$$y_1(k) = 0$$
$$y_2(k) = 1 \quad (2)$$
$$y_3(k) = 0$$

$$y_1(k) = 0$$
$$y_2(k) = 0 \quad (3)$$
$$y_3(k) = 1$$

donde (1) son para el objeto 1, (2) son para el objeto 2, y (3) son para el objeto 3.

### A. Sistema de inferencia difuso

El sistema de inferencia difuso aproxima las salidas reales $y_l(k)$ como sigue:

$$\hat{y}_l(k) = \frac{\sum_{j=1}^{m} v_{lj}(k) \alpha_j(x_j(k))}{\sum_{j=1}^{m} \alpha_j(x_j(k))} \quad (4)$$

donde $i=1,5$, $l=1,3$, $j=1,m$, $\hat{y}_l(k)$ son las salidas, $v_{lj}(k)$ son los centros de las funciones de membresía de las salidas. $m$ son el número de reglas. $\alpha$ son las funciones de Gauss dadas como sigue:

$$\alpha_j(x_j(k)) = \exp(x_j(k)) \quad (5)$$

donde $j=1,\ldots,m$, $x_j(k) = -\sum_{i=1}^{5} \frac{[z_i(k)-c_{ij}(k)]^2}{\sigma_{ij}^2(k)}$, $c_{ij}(k)$ y $\sigma_{ij}(k)$ son los centros y los anchos de las funciones de membresía de las entradas, $z_i(k)$ son las cinco entradas reales con diferentes valores para los tres objetos. La actualización de los parámetros de las funciones de membresía de las entradas y las salidas para el entrenamiento son como sigue:

$$v_{lj}(k+1) = v_{lj}(k) - \eta_{fs} \frac{\alpha_j(x_j(k))}{\sum_{j=1}^{m} \alpha_j(x_j(k))} e_l(k)$$

$$c_{ij}(k+1) = c_{ij}(k) - \eta_{fs} \beta_c(k) e_l(k)$$
$$\sigma_{ij}(k+1) = \sigma_{ij}(k) - \eta_{fs} \beta_\sigma(k) e_l(k) \quad (6)$$

$$\beta_c(k) = \frac{2\alpha_j(x_j(k))[z_i(k)-c_{ij}(k)][v_{lj}(k)-\hat{y}_l(k)]}{\left(\sum_{j=1}^{m} \alpha_j(x_j(k))\right) \sigma_{ij}^2(k)}$$

$$\beta_\sigma(k) = \frac{2\alpha_j(x_j(k))[z_i(k)-c_{ij}(k)]^2[v_{lj}(k)-\hat{y}_l(k)]}{\left(\sum_{j=1}^{m} \alpha_j(x_j(k))\right) \sigma_{ij}^3(k)}$$

donde $0<\eta_{fs}<1$ es la constante de la velocidad de aprendizaje, los parámetros iniciales $v_{lj}(1)$, $c_{ij}(1)$, y $\sigma_{ij}(1)$ se seleccionan como números aleatorios de 0 a 1, y $e_l(k)$ son los errores del algoritmo definidos como sigue:

$$e_l(k) = \hat{y}_l(k) - y_l(k) \quad (7)$$

La diferencia entre este algoritmo y el encontrado en la literatura es que el segundo usa el producto de las funciones de membresía de las entradas, mientras que el primero usa la sumatoria.

## V. ALGORITMO INTELIGENTE FINAL

En esta sección se explican los pasos para la programación en el aprendizaje del algoritmo propuesto.

### A. Función de inferencia difusa

La función de inferencia difusa es como sigue:

1) Obtener las salidas del sistema no lineal $y_l(k)$ con las ecuaciones (1), (2), y (3). El número de salidas $l=1,3$ se selecciona en relación al sistema no lineal.

2) Seleccionar los siguientes parámetros; $v_{lj}(1)$, $c_{ij}(1)$, y $\sigma_{ij}(1)$ son números aleatorios entre 0 y 1; $i=1,5$, $j=1,m$, $m$ como un número entero; obtener las salidas del sistema difuso como $\hat{y}_l(1)$ con las ecuaciones (4), (5).

3) Para cada iteración $k$, obtener las salidas del sistema difuso $\hat{y}_l(1)$ con las ecuaciones (4), (5), obtener el error del

algoritmo $e_l(k)$ con la ecuación (7), y actualizar los parámetros $v_{lj}(k+1)$, $c_{ij}(k+1)$, y $\sigma_{ij}(k+1)$ con la ecuación (6).

4) Notar que el algoritmo podría tener un mejor desempeño al cambiar los valores de $m$, $v_{lj}(1)$, $c_{ij}(1)$, o $\sigma_{ij}(1)$.

## VI. Método de clasificación con el sistema de inferencia difuso

El número total de datos se divide en dos partes, la primera es para el aprendizaje, y la segunda es para el examen.

### A. Entrenamiento

Los datos se dividen en tres partes, la primer parte se usa para el aprendizaje del comportamiento del objeto 1, la segunda parte se usa para el aprendizaje del comportamiento del objeto 2, y la tercera parte se usa para el aprendizaje de comportamiento del objeto 3.

### B. Examen

Los datos se dividen en 16 partes, para cada una de las partes, se usa el error cuadrático medio de raíz (ECMR) para la evaluación el cual es como sigue:

$$ECMR = \left( \frac{1}{T} \sum_{k=1}^{T} \sum_{l=1}^{3} e_l^2(k) \right)^{\frac{1}{2}} \quad (8)$$

donde $e_l(k)$ son los errores de (7) para el sistema de inferencia difuso, $T$ es el número de iteraciones, 3 es el número de salidas.

Si se satisface el siguiente criterio:

$$ECMR < \epsilon \quad (9)$$

entonces el parámetro llamado Examen Cierto (EC) se incrementa como sigue:

$$EC = EC + 1 \quad (10)$$

Después, el parámetro llamado Promedio de Exámenes Ciertos (PEC) se obtiene como sigue:

$$PEC = \frac{EC}{16} * 100 \quad (11)$$

donde 16 es el número total de exámenes.

Ahora, el número total de entradas también se divide en tres partes, la primera se emplea para el examen del comportamiento del objeto 1, la segunda parte se emplea para el examen de aprendizaje del comportamiento del objeto 2, y la tercer parte se emplea para el examen del aprendizaje del comportamiento del objeto 3.

En la primera de las tres partes, los datos de entrada del objeto 1 se utilizan para el examen, y los datos de salida para los tres objetos se utilizan para el examen. Entonces, el EC de (10) para el objeto 1 se debería incrementar, mientras que el EC de (10) para los otros objetos no se debería incrementar. Entonces, el ATT de (11) para el objeto 1 debería ser el más alto.

En la segunda de las tres partes, los datos de entrada del objeto 2 se utilizan para el examen, y los datos de salida para los tres objetos se utilizan para el examen. Entonces, el EC de (10) para el objeto 2 se debería incrementar, mientras que el EC de (10) para los otros objetos no se debería incrementar. Entonces, el ATT de (11) para el objeto 2 debería ser el más alto.

En la tercera de las tres partes, los datos de entrada del objeto 3 se utilizan para el examen, y los datos de salida para los tres objetos se utilizan para el examen. Entonces, el EC de (10) para el objeto 3 se debería incrementar, mientras que el EC de (10) para los otros objetos no se debería incrementar. Entonces, el ATT de (11) para el objeto 3 debería ser el más alto.

Los tres párrafos anteriores construyen la matriz de confusión para el examen de las concentraciones. El porcentaje de la eficiencia del sistema es el porcentaje de datos que se clasifican correctamente, y se obtiene como la sumatoria de los porcentajes de los términos de la diagonal principal en la matriz de confusión.

## VII. Resultados

En esta sección, dos ejemplos muestran la comparación entre el sistema difuso propuesto y el sistema difuso de [13] para la clasificación de tres objetos. El objetivo es que los datos se clasifiquen correctamente para los tres objetos: frutas para el ejemplo 1, y frutas o verduras para el ejemplo 2.

La Fig. 9 muestra la nariz electrónica de este estudio para obtener las concentraciones de los tres objetos de los ejemplos.

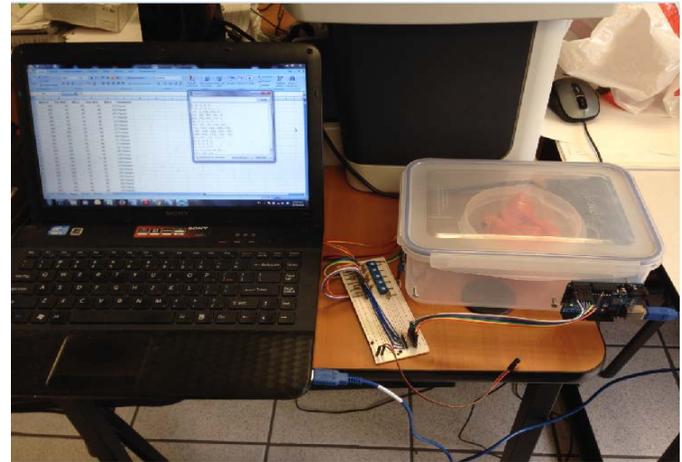

Figura 9. Nariz electrónica.

### A. Ejemplo 1

Esta sección presenta la matriz de confusión para el sistema de inferencia difuso en la clasificación de las concentraciones de los tres objetos: papaya como el objeto 1, naranja como el objeto 2, y manzana como el objeto 3. 400 iteraciones de datos se usan para el aprendizaje y 400 iteraciones de datos se usan para el examen.

*1) Sistema de inferencia difuso:* Se emplea el sistema de inferencia difuso de las ecuaciones (1)-(3), (4)-(7), (8)-(11) con parámetros $v_{lj}(1)$=rand, $c_{ij}(1)$=rand, $\sigma_{ij}(1)$=rand , $\epsilon = 1 \times 10^{-1}$. rand es un número aleatorio entre 0 y 1. La Tabla II muestra la matriz de confusión del sistema difuso de [13] con la nariz electrónica para m=20. La Tabla III muestra la matriz de confusión del sistema difuso propuesto con la nariz electrónica para m=10.

TABLA II. RESULTADOS DEL SISTEMA DE [13] EN EL EJEMPLO 1.

|  |   | $P$ | $N$ | $M$ |
|---|---|---|---|---|
|  | $P$ | 31.25 | 0 | 0 |
| Salidas | $N$ | 0 | 31.25 | 0 |
|  | $M$ | 0 | 0 | 37.5 |
|  |   | \multicolumn{3}{c}{Entradas} |

TABLA III. RESULTADOS DEL SISTEMA PROPUESTO EN EL EJEMPLO 1.

|  |   | $P$ | $N$ | $M$ |
|---|---|---|---|---|
|  | $P$ | 31.25 | 0 | 0 |
| Salidas | $N$ | 0 | 31.25 | 0 |
|  | $M$ | 0 | 0 | 37.5 |
|  |   | \multicolumn{3}{c}{Entradas} |

donde P es la papaya, N es la naranja, y M es la manzana.

*2) Descripción de los resultados experimentales:* De la Tabla II y la Tabla III, se puede observar que el sistema de inferencia difuso obtiene el mismo resultado al considerar el sistema difuso de [13] y el sistema difuso propuesto para la clasificación de las frutas debido a que las matrices de confusión de los sistemas difusos muestran una eficiencia del sistema de 100%, pero la diferencia radica en que el sistema de [13] usa 20 reglas, mientras que el sistema propuesto usa 10 reglas. Entonces, el sistema de inferencia difuso propuesto si es conveniente para la estrategia de clasificación propuesta en el caso de las frutas debido a que es más compacto.

*B. Ejemplo 2*

Esta sección presenta la matriz de confusión para el sistema de inferencia difuso en la clasificación de las concentraciones de los tres objetos: cebolla como el objeto 1, guayaba como el objeto 2, y plátano como el objeto 3. 400 iteraciones de datos se usan para el aprendizaje y 400 iteraciones de datos se usan para el examen.

*1) Sistema de inferencia difuso:* Se emplea el sistema de inferencia difuso de las ecuaciones (1)-(3), (4)-(7), (8)-(11) con parámetros $v_{lj}(1)$=rand, $c_{ij}(1)$=rand, $\sigma_{ij}(1)$=rand, $\in =1\times 10^{-1}$. rand es un número aleatorio entre 0 y 1. La Tabla IV muestra la matriz de confusión del sistema difuso de [13] con la nariz electrónica para m=20. La Tabla V muestra la matriz de confusión del sistema de inferencia difuso con la nariz electrónica para m=10.

TABLA IV. RESULTADOS DEL SISTEMA DE [13] EN EL EJEMPLO 2.

|  |   | $C$ | $G$ | $P$ |
|---|---|---|---|---|
|  | $C$ | 31.25 | 0 | 0 |
| Salidas | $G$ | 0 | 31.25 | 0 |
|  | $P$ | 0 | 0 | 37.5 |
|  |   | \multicolumn{3}{c}{Entradas} |

TABLA V. RESULTADOS DEL SISTEMA PROPUESTO EN EL EJEMPLO 2.

|  |   | $C$ | $G$ | $P$ |
|---|---|---|---|---|
|  | $C$ | 31.25 | 0 | 0 |
| Salidas | $G$ | 0 | 31.25 | 0 |
|  | $P$ | 0 | 0 | 37.5 |
|  |   | \multicolumn{3}{c}{Entradas} |

donde C es la cebolla, G es la guayaba, y P es el plátano.

*2) Descripción de los resultados experimentales:* De la Tabla IV y la Tabla V, se puede observar que el sistema de inferencia difuso obtiene el mismo resultado al considerar el sistema difuso de [13] y el sistema difuso propuesto para la clasificación de las frutas y los vegetales debido a que las matrices de confusión de los sistemas difusos muestran una eficiencia del sistema de 100%, pero la diferencia radica en que el sistema de [13] usa 20 reglas, mientras que el sistema propuesto usa 10 reglas. Entonces, el sistema de inferencia difuso propuesto si es conveniente para la estrategia de clasificación propuesta en el caso de las frutas y verduras debido a que es más compacto.

VIII. CONCLUSIÓN

En este artículo, se introdujo un método para la clasificación de frutas o verduras. La estrategia se ha probado con el sistema de inferencia difuso. El método propuesto produce una eficiencia del sistema de 100%. Entonces, el sistema de inferencia difuso si es conveniente para la estrategia de clasificación propuesta en el caso de las frutas o verduras. La técnica propuesta se puede aplicar para la clasificación de diferentes objetos como son las frutas, las verduras, los líquidos, las comidas, o la basura. Como futura investigación, éste método se podría combinar con los sistemas envolventes, o se podría aplicar para la clasificación de otros objetos.

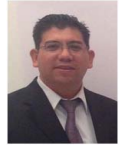
**Juan Manuel Stein** is currently a Ph. D. candidate in Computer Systems from the University DaVinci, he is a full time professor of the Tecnológico de Estudios Superiores de Ecatepec, he has published 2 papers in international journals, and his research is focused in human computer interfaces and pattern recognition.

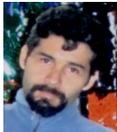
**José de Jesús Rubio (M'08)** is a full time professor of the Sección de Estudios de Posgrado e Investigación, ESIME Azcapotzalco, Instituto Politécnico Nacional. He has published 78 papers in International Journals, 1 International Book, 8 chapters in International Books, and he has presented 30 papers in International Conferences with 700 citations. He is a member of the IEEE AFS Adaptive Fuzzy Systems. He is member of the National Systems of Researchers with level II. He has been the tutor of 3 P.Ph.D. students, 5 Ph.D. students, 29 M.S. students, 4 S. students, and 17 B.S. students.

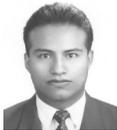
**Ramón Silva-Ortigoza** has been a Researcher at the Department of Mechatronics, CIDETEC, Instituto Politécnico Nacional, since 2006, and belongs to SNI-CONACYT, Mexico. He is a coauthor of the book Control Design Techniques in Power Electronics Devices (London, U.K.: Springer-Verlag, 2006). He has published more than 25 papers in credited journals. His research interests include control of mechatronic systems, mobile robotics, control in power electronics, and development of educational technology.

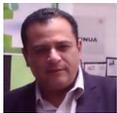
**Francisco Jacob Ávila** has the Ph. D. degree in Computer Systems from the Da Vinci University, currently serves as a research professor at the Tecnológico de Estudios Superiores de Ecatepec, he has published 2 papers in international journals, and his research is focused in artificial intelligence, pattern recognition, data mining, and embedded systems, among others.

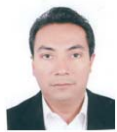
**Adolfo Meléndez** is currently a Ph. D. candidate in Computer Systems from the University DaVinci, he is full time professor at the Tecnológico de Estudios Superiores de Ecatepec, he has published 2 papers in international journals, and his research is focused in human interfaces computer and pattern recognition.